\documentclass[journal]{IEEEtran}
\usepackage{amsmath,amsfonts}
\usepackage{amssymb}
\usepackage{algorithmic}
\usepackage{algorithm}
\usepackage{array}
\usepackage[caption=true,font=normalsize,labelfont=sf,textfont=sf]{subfig}
\usepackage{textcomp}
\usepackage{stfloats}
\usepackage{url}
\usepackage{verbatim}
\usepackage{graphicx}
\usepackage{cite}
\usepackage{epstopdf}
\usepackage{accents}

\begin{document}

\title{WDMoE: Wireless Distributed Mixture of Experts for Large Language Models}

\author{Nan Xue, Yaping Sun, Zhiyong Chen, Meixia Tao, \emph{Fellow, IEEE}, Xiaodong Xu, Liang Qian, \\ Shuguang Cui, \emph{Fellow, IEEE}, Wenjun Zhang,  \emph{Fellow, IEEE}, and Ping Zhang,  \emph{Fellow, IEEE} 
\thanks{N. Xue, Z. Chen, M. Tao, L. Qian and W. Zhang are with the Cooperative Medianet Innovation
Center, Shanghai Jiao Tong University, Shanghai 200240, China. (email:
\{nan.xue, zhiyongchen, mxtao, lqian, zhangwenjun\}@sjtu.edu.cn)

Y. Sun is with the Department of Broadband Communication, Pengcheng
Laboratory, Shenzhen 518000, China. Y. Sun is also
with the Future Network of Intelligent Institute (FNii), the Chinese University
of Hong Kong (Shenzhen), Shenzhen 518172, China (email: sunyp@pcl.ac.cn).

X. Xu and P. Zhang are with the Beijing University of Posts and Telecommunications, Beijing 100876, China, and affiliated with the Department of
Broadband Communication, Peng Cheng Laboratory, Shenzhen 518000, China
(email: xuxd@pcl.ac.cn; pzhang@bupt.edu.cn).

S. Cui is with the School of Science and Engineering (SSE) and the
Future Network of Intelligent Institute (FNii), the Chinese University of Hong
Kong (Shenzhen), Shenzhen 518172, China. S. Cui is also affiliated with the
Department of Broadband Communication, Peng Cheng Laboratory, Shenzhen
518000, China (email: shuguangcui@cuhk.edu.cn).}
\thanks{Part of this manuscript will be presented at IEEE GLOBECOM 2024 \cite{xue2024wdmoe}.}}

\markboth{}%
{Shell \MakeLowercase{\textit{et al.}}: A Sample Article Using IEEEtran.cls for IEEE Journals}


\maketitle

\begin{abstract}
Large Language Models (LLMs) have achieved significant success in various natural language processing tasks, but the role of wireless networks in supporting LLMs has not been thoroughly explored. In this paper, we propose a wireless distributed Mixture of Experts (WDMoE) architecture to enable collaborative deployment of LLMs across edge servers at the base station (BS) and mobile devices in wireless networks. Specifically, we decompose the MoE layer in LLMs by placing the gating network and the preceding neural network layer at BS, while distributing the expert networks among the devices. This deployment leverages the parallel inference capabilities of expert networks on mobile devices, effectively utilizing the limited computing and caching resources of these devices. Accordingly, we develop a performance metric for WDMoE-based LLMs, which accounts for both model capability and latency. To minimize the latency while maintaining accuracy, we jointly optimize expert selection and bandwidth allocation based on the performance metric. Moreover, we build a hardware testbed using NVIDIA Jetson kits to validate the effectiveness of WDMoE. Both theoretical simulations and practical hardware experiments demonstrate that the proposed method can significantly reduce the latency without compromising LLM performance.  
\end{abstract}

\begin{IEEEkeywords}
  Distributed Large Language Models, Mixture of Experts, Expert Selection, Wireless Communications, Resource Allocation
\end{IEEEkeywords}

\section{Introduction}
The exciting advancements in large language models (LLMs) have sparked a new wave of AI innovation. LLMs, exemplified by ChatGPT\cite{openai2022chatgpt}, have demonstrated emergent abilities\cite{wei2022emergent}, including better generalization, nuanced meaning comprehension, and remarkable reasoning and generation capabilities. These advancements have led to widespread applications across various fields, illuminating the vision of artificial general intelligence (AGI)\cite{bubeck2023sparks}. In the field of 6G wireless networks, LLMs have been used for wireless network resource allocation\cite{shao2024wirelessllm, lee2024llm, 10582827}, and applied in internet of vehicles\cite{zhou2024large} and immersive communications\cite{sehad2024generative}.

The emergent abilities of LLMs stem from extensive computation, a large number of model parameters, and massive training datasets\cite{wei2022emergent, kaplan2020scaling, hoffmann2022training}. The vast number of model parameters poses significant challenges for training, inference, and deployment. The training phase of LLMs involves significant costs in time and computational power for most individuals and organizations. Regarding LLMs inference and deployment, they also require fast responses and ample memory. In this paper, we mainly focus on LLMs inference and deployment.

Currently, LLMs can be classified into cloud-based LLMs and on-device LLMs based on their deployment characteristics. Cloud servers with numerous graphics processing units (GPUs) and sufficient power supply are responsible for the majority of model inference and deployment. Due to concerns over latency and data privacy, the potential of on-device LLMs is gaining increasing attention\cite{xu2024device}. Researchers compress LLMs through pruning\cite{ma2023llm}, quantization\cite{lin2024awq}, and distillation\cite{abdin2024phi} to meet the memory, computation, and energy requirements of mobile devices. Limited by generation speed and model capabilities, even a company as strong as Apple has not been able to deploy a fully satisfactory LLM on mobile devices. On the latest iPhone 16 Pro series, only simple tasks are completed locally by a model with around 3 billion parameters, whereas complex tasks are still handled by cloud-based models like ChatGPT\cite{gunter2024apple, apple2024chat}. Although in practical LLMs application transformer's KV cache\cite{pope2023efficiently} can speed up the inference, it will cause considerable memory overhead, which presents an obstacle for on-device LLMs.

In light of the rapid advancements of LLMs and the widespread adoption of 5G/6G wireless networks, a natural question arises: \textbf{can wireless networks support LLMs? If so, how?} The answer lies in fully leveraging the multidimensional resources of wireless networks, incorporating computing, communications, and caching (3C) to support LLMs and enhance user experience \cite{primary3C}. As a key technology of 5G, mobile edge computing (MEC) has been thoroughly studied to improve the quality of service for various network applications. Recently, edge-cloud collaborative training and fine-tuning are also researched\cite{10577141, 10628022}. However, there is a lack of specialized optimization research tailored to the characteristics of LLMs in the scenario of distributed deployment in wireless networks. To address this issue, this paper aims to bridge the gap by proposing a distributed deployment of LLMs powered by mixture of experts (MoE), modeling and optimizing the latency during the inference phase, which can efficiently utilize the 3C resources at both the MEC server and mobile devices.

The core idea of MoE architecture is to sparsify the neural networks (NNs) based on the observation of sparse activation of NNs and allow each expert to specialize in their specific tasks, reducing the floating point operations (FLOPs) to achieve acceleration and increasing model capacity\cite{shazeer2017outrageously}. Transformer with MoE\footnote{In this paper, MoE refers to Transformer with MoE unless otherwise specified.}\cite{lepikhin2020gshard} replaces the original single dense feedforward neural network (FFN) with a gating network and multiple smaller expert networks, which often have the same structure and different hidden layer dimensions. The expert networks within the same MoE layer operate in parallel and independently, providing high fault tolerance and making it well-suited for wireless distributed deployment. Besides, we find during the inference phase of MoE-based LLMs, decreasing the number of participating experts will not degrade model performance, showcasing the robustness of MoE. Capitalizing on these attributes, we can deploy the gating network at MEC server and each expert network in an MoE layer on diverse mobile devices. Furthermore, due to the diversity of computing capabilities among the mobile devices and communication qualities between  the base station (BS) and the mobile devices, how to achieve a good balance between the inference accuracy and latency also requires careful consideration.

\subsection{Related Work}
Current mainstream LLMs utilize Transformer architecture\cite{vaswani2017attention}. LLMs based on the Transformer architecture can be categorized into three types: encoder-decoder, decoder-only, and encoder-only. For encoder-decoder structure, in the vanilla Transformer\cite{vaswani2017attention}, the multi-head attention mechanism fully replaces traditional recurrent neural networks, making the architecture more suitable for parallel computing. For decoder-only structure, in \cite{radford2018improving}, a decoder-only Transformer is utilized with the optimization objective of standard language modeling. This work achieves state of the art performance, and the model is named GPT. For the encoder-only structure, \cite{devlin2018bert} designs Bidirectional Encoder Representations from Transformers (BERT) and applies the training paradigm of pre-training and fine-tuning to language model. BERT includes a base model with 110 million parameters, similar to GPT and a large model with 340 million parameters. The work \cite{radford2018improving} and \cite{devlin2018bert} lay the foundation for the model architecture, training paradigm, and development direction of large language models. GPT-2 is a scaled-up version of GPT with 1.5 billion parameters, focusing on zero-shot performance in \cite{radford2019language}. The work \cite{kaplan2020scaling} proposes the scaling law of transformer-based language models and suggests that larger models trained on more data tend to perform better, supporting the later development of large language models. GPT-3 is released in \cite{brown2020language} with 175 billion parameters and was the largest language model at that time. It is applied without fine-tuning and achieves state-of-the-art performance on various benchmarks. Following the  scaling law, an increasing number of large language models have appeared with more parameters, and the largest open source model is Llama 3.1, with 405 billion parameters, based on a dense Transformer\cite{dubey2024llama}. 

Large model size makes the training and inference computationally expensive. Various methods of training and inference have been proposed to improve model performance without excessive computational costs\cite{shazeer2017outrageously}, among which the MoE architecture has been well-researched. The basic MoE layer consists of a gating network and a number of experts, which are simple feed-forward neural networks\cite{jacobs1991adaptive}. The work \cite{shazeer2017outrageously} introduces a Sparsely-Gated MoE and expands the LSTM-based language model to a configuration with 137 billion parameters. In the Transformer era, \cite{lepikhin2020gshard} replaces all Feed-forward layers in the model with MoE layers and achieves enhanced model capacity while substantially reducing the training time. Mistral AI successively releases two open-source MoE large language models, Mixtral-8x7B and Mixtral-8x22B, both achieving the state-of-the-art performance among open-source large models at the time of their release\cite{jiang2024mixtral}. The work \cite{chen2023adamv} introduces an adaptive MoE framework, AdaMV-MoE, designed for multi-task vision recognition. Unlike conventional MoE approaches with a fixed number of experts, AdaMV-MoE dynamically adjusts the number of active experts per task based on training dynamics, eliminating the need for manual tuning of model capacity. Based on the idea that harder tasks need more experts, the work \cite{huang2024harder} develops a method for dynamically adjusting the number and selection of experts to reduce computational costs for simple tasks while enhancing model performance for complex ones. Advance in MoE models have unlocked the potential of model capability and computational efficiency. 

On-device LLMs have been extensively researched as a promising solution for personal assistants. By running LLMs directly on devices such as smartphones or edge servers, they offer enhanced privacy, lower latency, and reduced dependence on cloud servers. The work \cite{yin2024llm} devises a KV cache compression and swapping method with accuracy tolerance awareness, significantly reducing the context switching latency and extending the capacity of active contexts. In \cite{yu2024edge}, a highly efficient computation and memory framework named Edge-LLM is proposed, reducing computation and memory overhead through layer-wise unified compression and adaptive backpropagation depth reduction. MobileLLM is a model family with sub-billion parameters that surpasses previous sub-billion models on mainstream benchmarks by adopting deep and thin network architectures, embedding sharing, and a grouped-query attention mechanism\cite{liu2024mobilellm}. With privacy considerations, \cite{peng2024pocketllm} employs derivative-free optimization to update the local LLM's parameters during on-device fine-tuning phase under the resource-constrained circumstances.
\begin{figure*}[t]
	\centering
	\subfloat[]{\includegraphics[scale=0.37]{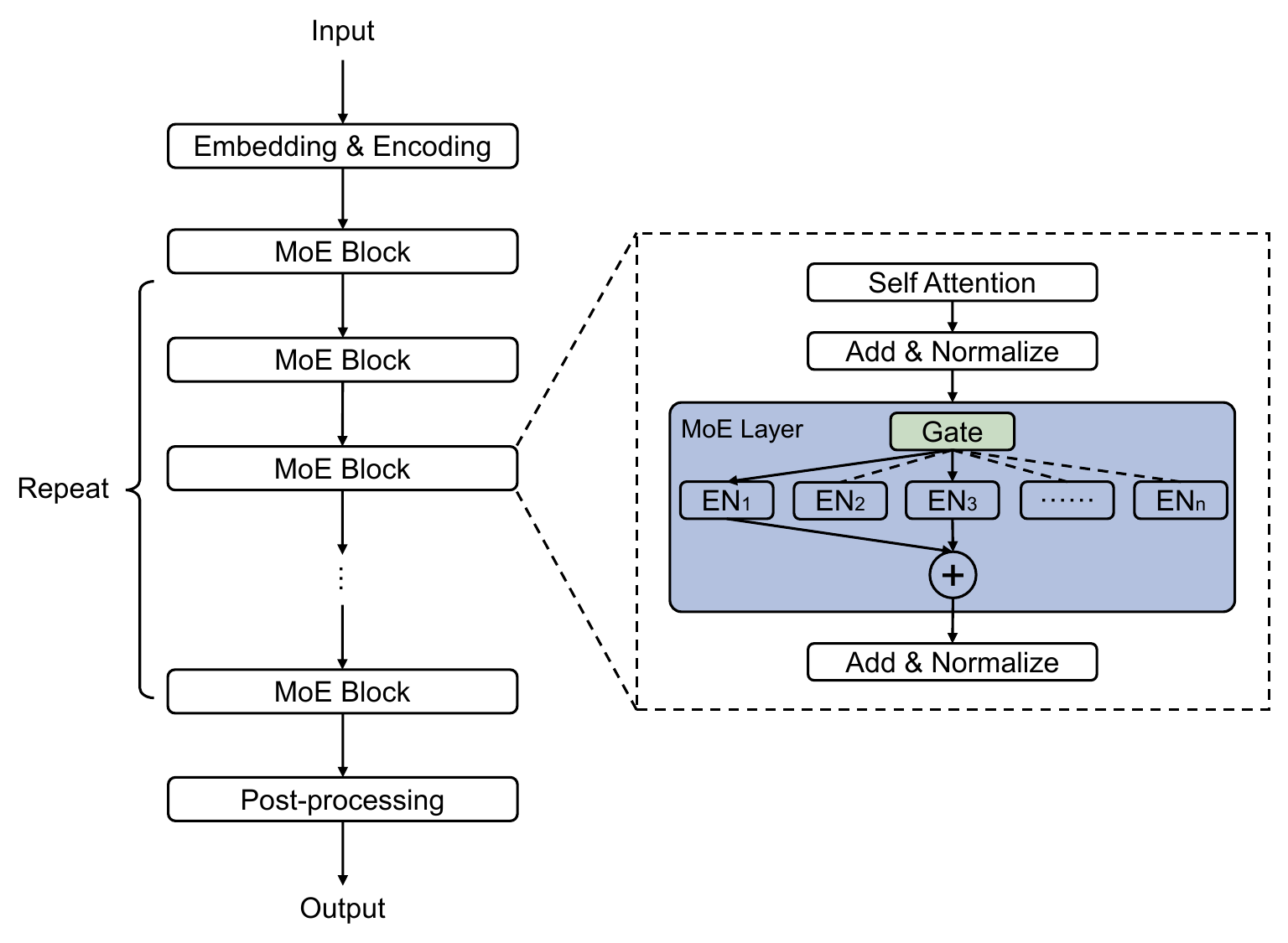}\label{moe}}
	\subfloat[]{\includegraphics[scale=0.37]{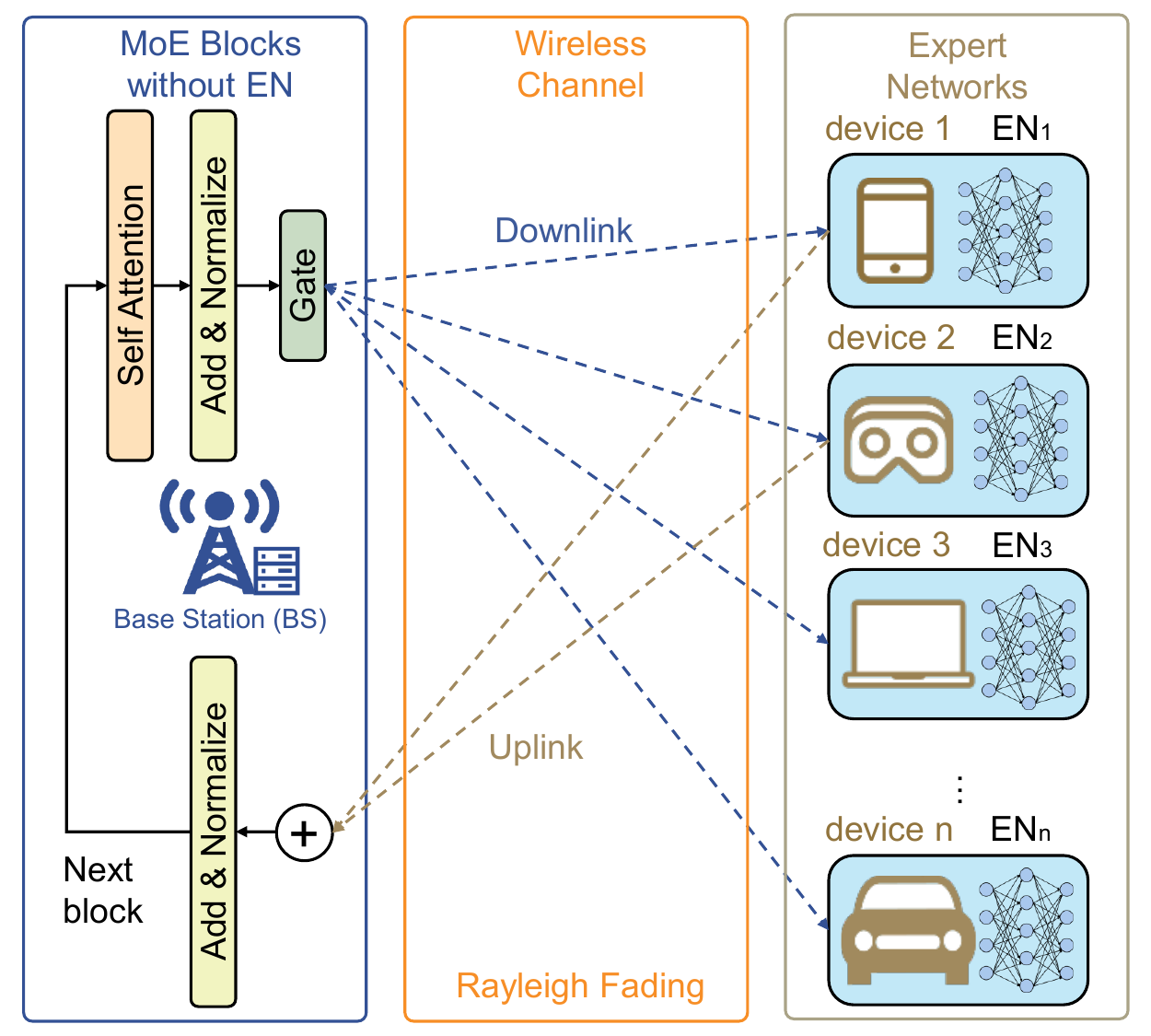}\label{sys}}\\
	\caption{(a) MoE-based LLMs architecture\cite{lepikhin2020gshard}; (b) The proposed WDMoE-based LLMs system model.}
	\label{mas}
\end{figure*}

\subsection{Contributions}
Motivated by the above, we explore a wireless distributed deployment architecture for LLMs with MoE, define and model key performance metrics of LLMs within the framework, and optimize resource allocation and expert selection to improve the LLMs service capabilities within the wireless network. The main contributions of this paper are summarized as follows:

\begin{itemize}
	\item \textbf{We propose a novel wireless distributed MoE architecture, WDMoE, for LLMs}. The fundamental principle of the WDMoE architecture is to leverage the independence and parallelism of expert networks by deploying them on multiple mobile devices, thereby utilizing the resource advantages of wireless networks to facilitate distributed deployment of large models. This architecture places the computationally intensive and memory-demanding multi-head attention mechanism on the MEC servers at BS. Through collaborative token processing between BS and mobile devices, the wireless network can support a large number of LLMs service requests, addressing the challenges of limited memory on a single device and data privacy concerns in cloud-based LLMs deployment.
	\item \textbf{We establish latency-aware performance metrics of the WDMoE-based LLMs.} By analyzing the attention mechanism, we find that, during the wireless distributed deployment of LLMs with MoE, the latency for expert networks on different mobile devices to process tokens and return them to the BS varies. This discrepancy can cause earlier-arriving tokens to wait at the attention module. We model this latency, referred to as attention waiting latency, and design a bilevel optimization problem where the upper-level objective is to minimize this latency while ensuring model capabilities at the lower level, thereby reducing  latency without compromising model performance. In addition, we propose an innovative metric, the weight latency ratio (WLR), to comprehensively consider the output weights of the MoE gating network and the attention waiting latency of each device. 
\item \textbf{We develop an expert selection policy to improve the performance}. Based on the defined WLR, the expert selection policy can process tokens by considering the processing time and weight of each mobile device for allocated tokens. It dynamically adjusts the number of experts per token, thereby reducing network transmission and computational load.
	\item \textbf{We build a hardware testbed using NVIDIA Jetson kits to validate the effectiveness of the proposed WDMoE.} We use three NVIDIA Jetson kits and one personal computer (PC) with an NVIDIA RTX 4070 Ti, serving as four mobile devices, each running one expert for each layer, and communicating with the server via WiFi. Both simulation results and hardware experiment results show that the proposed WDMoE architecture and expert selection policy effectively reduce the latency experienced by users. For example, compared to vanilla expert selection, the proposed WDMoE reduces latency by 45.75\% on average on the PIQA dataset without model capability deterioration.

\end{itemize}

The rest of this paper is organized as follows. The WDMoE-based LLM architecture is introduced in Section II. The system model and problem formulation are presented in Section III. The WDMoE expert selection policy and bandwidth allocation algorithm are introduced in Sections IV. Section V shows extensive simulation results. Finally, the hardware testbed experiments are presented in Section VI, and conclusions are drawn in Section VII.

\section{The Proposed WDMoE Based LLM}
\subsection{Structure of MoE-based LLMs}
The network structure of MoE-based LLMs is depicted in Fig. \ref{mas}\subref{moe}, where each MoE block is based on a Transformer. In these blocks, the FFN module is replaced by an MoE layer\cite{jiang2024mixtral}. An MoE layer consists of a gating network, also known as a router, and multiple expert networks, each of which can be any type of neural network. The gating network, which is a simple neural network, processes the input token to produce a weight vector for each expert.

The number of expert networks is denoted as $n$. Let $\mathbf{x}^i \in \mathbb{R}^{J \times m}$ be the input token set of the gating network in the $i$-th MoE block, where $J$ is the total number of input tokens of all prompts at present, and $m$ is the dimension of a token's embedding. $I$ denotes the total number of MoE blocks in the LLMs. The element $x^{i}_{j} \in \mathbf{x}^i$ denotes the $j$-th input token of the $i$-th MoE block. This token $x^i_j$ serves as the input to the gating network, which then outputs weights $\mathbf{w}^{i}_{j}$ for the $j$-th token in the $i$-th block. Here, $\mathbf{w}^{i}_{j} \in \mathbb{R}^{n}$ is an $n$-dimension column vector representing weights allocated to $n$ experts. $x^i_j$ is also fed into the expert networks, whose output $\mathbf{y}^i_j$ is an $n$-dimension column vector. The output $o^i_j$ of the $j$-th token through the $i$-th MoE layer is:
\begin{equation}
  \label{eq:1}
  o^{i}_{j} = {\mathbf{w}^i_j}^\top {\mathbf{y}}^{i}_{j}.
  \end{equation}

\subsection{Distributed Deployment of WDMoE}
Experts operate in parallel and do not impact each other, making this approach suitable for deployment in distributed mobile edge networks. We consider a BS equipped with an edge server that possess powerful computing capability, to process multiple data streams simultaneously for $n$ mobile devices, as shown in Fig. \ref{mas}\subref{sys}. In the WDMoE architecture, the attention mechanism and gating network are deployed at the BS, while only expert networks are assigned to mobile devices. Typically, the expert network consists of a simple multilayer perceptron (MLP). When a user sends a prompt, its embedding operation can be completed either locally or at BS, depending on the choice of the user. In this paper, we focus on the communications and computing costs incurred during interactions between BS and mobile devices following the first embedding module. The proposed WDMoE can be integrated with existing splitting methods.

The mobile devices are connected to the BS via wireless links. We denote the set of mobile devices as $\mathcal{U} \triangleq \{1, \cdots, U\}$. Let $B_k \in [0, B]$ denote the bandwidth allocated to the $k$-th device. For the downlink transmission, the data transmission rate from the BS to the $k$-th device is formulated as:
\begin{equation}
    {R}^{d}_{k} = B_{k} \log_2 \left(1 + \frac{{P}^{d}_{k} g_{BS,k}}{N_0 B_{k}}\right),\ \forall k \in \mathcal{U},
    \label{comm_down}
\end{equation}
where ${P}^{d}_{k}$ denotes the transmission power for device $k$, $g_{BS,k}$ represents the channel gain from the BS to the $k$-th device, and $N_0$ is the noise power spectral density. The uplink transmission rate from the $k$-th device to the BS is:
\begin{equation}
    {R}^{u}_{k} = B_{k} \log_2 \left(1 + \frac{{P}^{u}_{k}g_{k,BS}}{N_0 B_{k}}\right),\ \forall k \in \mathcal{U},
    \label{comm_up}
\end{equation}
where ${P}^{u}_{k}$ denotes the transmission power of the $k$-th device, and $g_{k,BS}$ represents the channel gain from the $k$-th device to the BS.
\begin{figure}
	\centering
	\includegraphics[width=1\columnwidth]{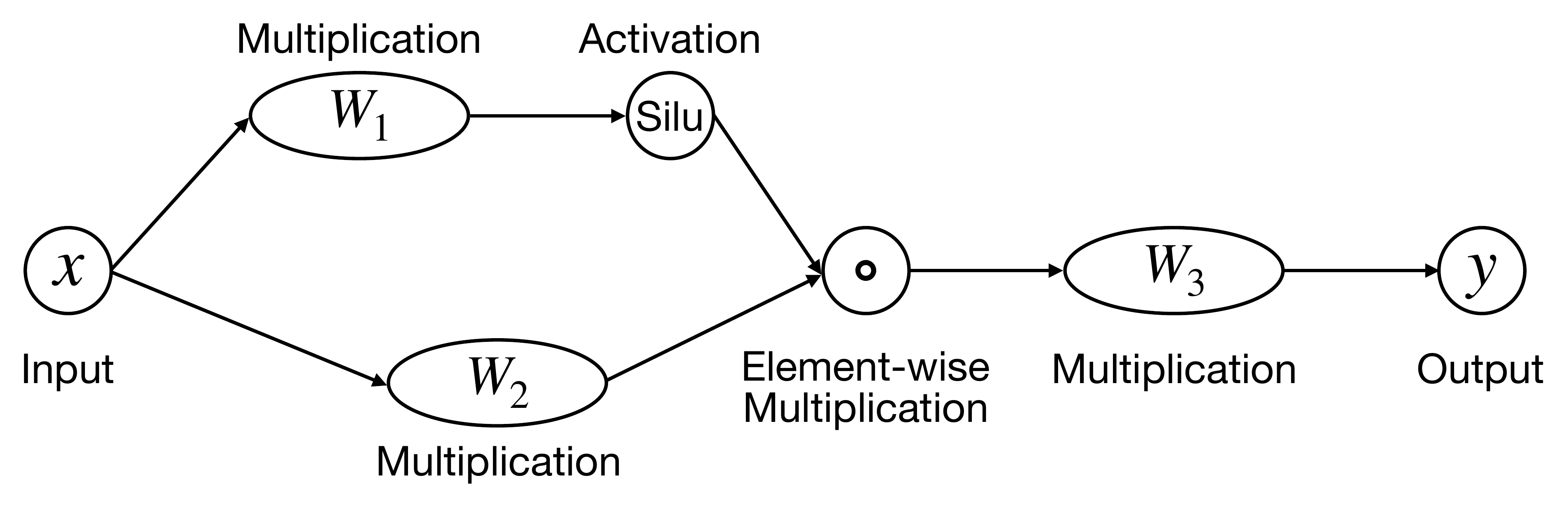}
	\caption{Expert network structure.}
	\label{FFN}
\end{figure}
For WDMoE, the token embeddings are transmitted between the BS and the devices. The size of each token embedding is denoted by $m$. The data size of a token embedding, represented as $L^{comm}$, can be calculated as follows:
\begin{equation}
    L^{comm} = \epsilon \times m,
\end{equation}
where $\epsilon$ is a coefficient determined by the quantization precision. For instance, a half-precision floating-point occupies 16 bits\cite{8766229} and thus $\epsilon = 16$. 

\begin{figure*}[t]
	\centerline{\includegraphics[scale=.5]{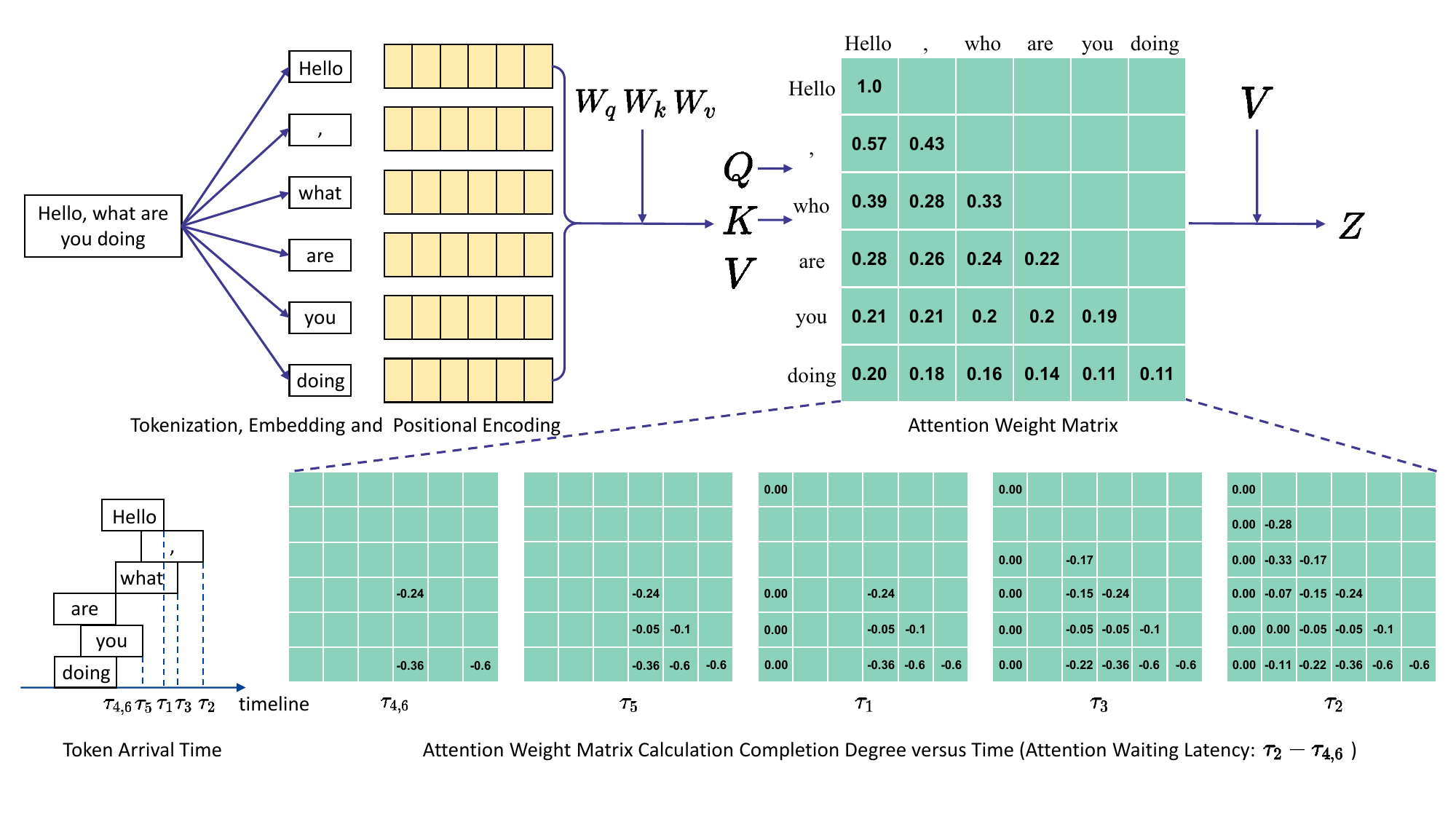}}
	\caption{An example of waiting in the attention mechanism.}
	\label{figure_attn}
\end{figure*}

We consider each device is equipped with at least one graphics processing unit (GPU). The expert network used in this paper is an FFN, comprising neural network linear layers, activation functions, and element-wise multiplications, as shown in Fig. \ref{FFN}. The primary floating-point operations include matrix multiplications and element-wise multiplications, as activation functions is expressed using these two fundamental operations. The FLOPs required by our expert network is given by \cite{molchanov2017pruning}:
\begin{equation}
    L^{comp} = 4 m \times m_{h} + 2 m_{h} \times m + \eta \times m_{h} + m_{h},
\end{equation}
where $m_h$ is the hidden dimension of the FFN and $\eta$ denotes the FLOPs required by the activation function in the network. Considering the strong computational capability of MEC servers, the computing latency at MEC server is negligible compared to the communications time and computation latency at mobile devices in this paper.

\section{Performance Metrics and Problem Formulation}

\subsection{Token Processing Latency}
The output tensor retains the same shape as the input tensor, indicating that the data transmission size for the uplink is equivalent to that for the downlink. The communication latency for the $j$-th token in the $i$-th block, processed by expert $k$, is defined as:

\begin{equation}
    t^{comm}_{i, j, k} = \frac{L^{comm}}{{R}^{d}_{k}} + \frac{L^{comm}}{{R}^{u}_{k}}.
    \label{t_comm}
\end{equation}

The computation latency for the $j$-th token in the $i$-th block processed by the expert $k$ is given by:
\begin{equation}
    t^{comp}_{i, j, k} = \frac{L^{comp}}{C_{k}},
    \label{t_comp}
\end{equation}
where $C_k$ denotes the computational capacity of expert $k$.

The total delay of the $j$-th token in the $i$-th block processed by the expert $k$ is given by:
\begin{equation}
    t_{i, j, k} = t^{comm}_{i, j, k} + t^{comp}_{i, j, k}.
    \label{t_total}
\end{equation}

\subsection{Attention Waiting Latency}
Once each expert completes processing a token, the result is transmitted back to the BS. The BS aggregates the results, after which the tokens are forwarded to the next block. The first operation in the subsequent block is the self-attention mechanism. To more accurately model the latency experienced by users, we account for the impact of token latency on the following block's modules. Although computation latency at the BS is disregarded, the latency of preceding tokens affects the overall practical latency. The attention mechanism processes all tokens as a sequence to compute attention scores and reconstructs the output tensor, which maintains the same shape as the input token. 

As illustrated in Fig. \ref{figure_attn}, the attention mechanism computes the attention weight matrix, containing the attention scores for each pair of input tokens. In the distributed deployment of WDMoE, all tokens are transmitted to various mobile devices, and their computation results arrive at the edge server at different times due to factors such as channel uncertainty, variations in mobile device workloads, and available resources. In Fig. \ref{figure_attn}, the token \texttt{doing} is the last token in the sequence but is among the first to be transmitted back to the server, along with the fourth token, \texttt{are}, at time $\tau_{4, 6}$. At this point, only the attention scores for these two token pairs can be calculated. The first two arriving tokens must wait for the remaining tokens to arrive. At time $\tau_2$, the entire masked attention weight matrix—formed by the multiplication of the query and key—is completed and used for weighted calculation with the corresponding values.


We denote the expert selection matrix for the $i$-th block as $\mathbf{Q}^i \in \mathbb{R}^{J \times U} $. Each element of $\mathbf{Q}^i$, denoted by $q^i_{j, k} \in \{0, 1\}$, represents the expert selected for the $j$-th token in the $i$-th block. All the expert selections $\mathbf{Q}^i, \forall i \in \{1, \cdots, I\}$ are concatenated to get $\mathbf{Q}$. Specifically, $q^i_{j,k} = 1$ indicates that the $k$-th expert is selected for the $j$-th token in the $i$-th block. The total delay of the sequence is determined by the device with the slowest processing and transmission time, necessitating load balancing. The total number of tokens in the $i$-th block assigned to the $k$-th device is given by:
\begin{equation}
	q^i_k = \sum_{j=1}^{J} q^i_{j,k}.
\end{equation}

Since all tokens have the same data size and computation load, we have $t_{i,k} = t_{i,j,k}$ for all $j$, and the total time required for device $k$ to process all allocated tokens is given by:
\begin{equation}
	t^i_k = q^i_k t_{i,k}.
\end{equation}

Finally, the actual latency from the gating network of the $i$-th block to the attention mechanism of the $(i+1)$-th block is:
\begin{equation}
	t^i = \max_{k \in \mathcal{U}} t^i_k.
\end{equation}

\subsection{Weight-to-Latency Ratio}
The attention waiting latency provides a more comprehensive and precise reflection of the latency experienced by users. For any device involved in the network, it is crucial to prioritize the overall benefits derived from processing the assigned tokens. Thus, we introduce a weight-to-latency ratio from the perspective of the processing device to quantify its processing efficiency. The WLR for device $k$ in the $i$-th block is defined as
\begin{equation}
    WLR^i_{k} = \frac{\sum_{j=1}^J q^i_{j,k} w^i_{j,k}}{t^i_k}.
    \label{WLR_eq}
\end{equation}
\subsection{Problem Formulation}
We formulate a bilevel optimization problem \cite{dempe2002foundations} to minimize the total system latency while ensuring comprehensive system performance. For the bilevel optimization, the upper level problem seeks to minimize total latency by optimizing the bandwidth allocation $\{B_k\}$ and the expert selection matrix $\mathbf{Q}$, whereas the lower level problem aims to maximize device profit by optimizing the expert selection matrix $\mathbf{Q}$. The upper-level problem is formulated as: 

\begin{align}
    &\llap{\text{P1:}} & \min_{\mathbf{B}, \mathbf{Q}} \quad & \sum_{i=1}^{I} t^i \nonumber \\
    && \text{s.t.} \quad & \sum_{k=1}^{U} B_k = B,  \label{b1}\\
    &&& B_k \geq 0, \ \forall k \in \mathcal{U}, \label{b2}\\
    &&& \mathbf{Q} \in \Phi(\mathbf{Q}), 
\end{align}
where $\Phi(\mathbf{Q}) = \arg max_{\mathbf{Q}} \sum_{i=1}^{I} \sum_{k=1}^{U} WLR^i_{k}$ is the expert selection solution space at the lower level problem. The constraints (\ref{b1}) and (\ref{b2}) represent that the bandwidth allocated to each device is within the total bandwidth $B$.

 The lower level problem is formulated as:
\begin{align}
    &\llap{\text{P2:}} &\max_{\mathbf{Q}} \quad & \sum_{i=1}^{I} \sum_{k=1}^{U} \frac{\sum_{j=1}^J q^i_{j,k} w^i_{j,k}}{q^i_k \cdot t_{i,k}} \nonumber \\
    && \text{s.t.} \quad & \sum_{k}^{K} q^i_{j,k} \geq 1, \ \forall j \in \{1, \cdots, J\},\\
    &&& q_{j,k}^i \in \{0,1\}.
\end{align}

\begin{figure*}[t]
	\centerline{\includegraphics[scale=0.5]{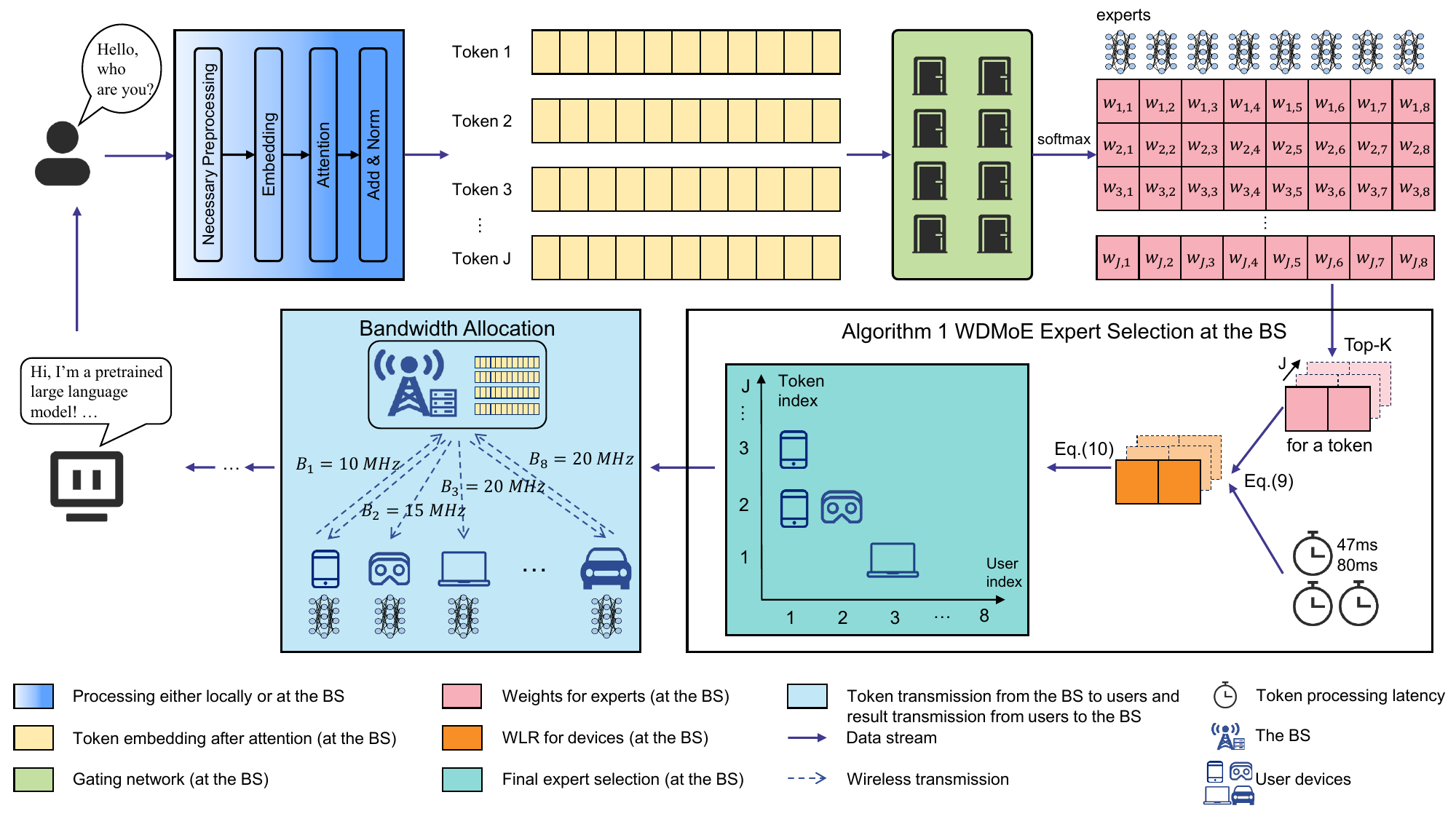}}
	\caption{The data stream, expert selection and bandwidth allocation in WDMoE. When a user sends a prompt to WDMoE, it is preprocessed, embedded, and subjected to attention operations either locally or at BS, depending on user preference. Each token's embedding is analyzed by a gating network to assign weights to each expert. WDMoE dynamically adjusts these weights, the number of experts, and optimize the bandwidth allocation based on gating network output and wireless channel conditions. The token embeddings are sent to the respective devices for processing by expert networks. Once processed, the results are sent back to BS, where they are weighted, summed, and then transformed from embeddings into words.}
	\label{figure_mechan}
\end{figure*}
The $j$-th token in $i$-th MoE block is either processed by the device $k$ or not, and each token is processed on at least one device. The lower level problem P2 is a combinatorial optimization problem and can be reduced to the set cover problem. Problem P2 is NP-hard, and thus the upper level problem P1 also becomes an NP-hard optimization problem.

\section{Expert Selection Policy and Bandwidth Allocation}
In this section, we first propose an expert selection policy to solve the lower level problem P2, based on which we optimize the bandwidth allocation to address problem P1. In particular, the expert selection policy is first designed to dynamically adjust the expert selection to improve the system's overall performance. Based on the proposed expert selection policy, the bandwidth allocation algorithm is devised to optimize the bandwidth allocation to minimize the total latency of the system. 

\subsection{Expert Selection Policy}
Retraining a LLM to account for wireless communication conditions presents significant challenges, particularly in enabling the model to recognize and adapt to varying channel conditions in distributed environments. Retraining the gating network of an MoE layer is generally not advisable, as it is both time-consuming and computationally demanding. Furthermore, integrating wireless communication factors may unintentionally disrupt the gating network, turning what is originally a single-objective optimization problem into a multi-objective problem, which can hinder model convergence. However, MoE-based LLMs are highly robust, even when expert selection deviates from the trained gating network’s outputs. In fact, moderate adjustments to expert selection are often tolerated and may sometimes even outperform the original selection. To address this, we propose a training-free scheme to adjust expert selection and effectively solve the lower-level optimization problem.

The data flow and the solution algorithm for the lower-level optimization problem are depicted in Fig. \ref{figure_mechan}. The process begins when a sequence of tokens is fed into the gating network, which assigns weights to each expert in the MoE layer. Since the parameters of the router are frozen, the output weights closely align with the model’s performance. The device latency is estimated at the BS, where the BS has access to the channel conditions of all devices and computes the latency based on \eqref{t_total}. Some devices may be located far from the BS or in areas with poor coverage, potentially leading to significant latency. The lower-level optimization objective is designed to reduce the workload without degrading model accuracy. To address this, we first compute the latency vector for each token as $\mathbf{t}^i_j = [t^i_{j,1}, \cdots, t^i_{j,k}]$, assuming bandwidth is evenly distributed. Next, cosine similarity is employed to assess the relationship between the weights and the latency. The cosine similarity is calculated as:
\begin{equation} 
{S}({\mathbf{w}^i_j}, \mathbf{t}^i_j) = \frac{{\mathbf{w}^i_j}^\top \mathbf{t}^i_j}{|\mathbf{w}^i_j| \cdot |\mathbf{t}^i_j|}, \label{sim} 
\end{equation}
where ${S}(\mathbf{w}^i_j, \mathbf{t}^i_j)$ represents the cosine similarity of the $j$-th token in the $i$-th block.

The expert selection policy is designed to adjust expert selection based on cosine similarity. We define a threshold $\theta$ to determine whether to drop the expert with minimal weights after Top-K selection. The experts are sorted from the highest weights to the lowest, and the top $k$ experts are selected. When ${S}(\mathbf{w}^i_j, \mathbf{t}^i_j) \leq \theta$, the expert with the lowest weights will be dropped. Through this mechanism, we can reduce communication and computational load without significant impact on model performance. If we increase the threshold, fewer experts will be chosen, which will decrease the total load, though it may also affect model performance with greater probability. This expert selection policy will benefit resource allocation in the WDMoE-based LLM due to the reduction in total load.

The threshold for expert selection can be adjusted based on the WLR, with a higher threshold leading to the exclusion of more experts and a corresponding increase in WLR to some extent. It is crucial to note that a moderate increase in the threshold can reduce latency without significantly impacting model performance. However, an excessive increase in the threshold may result in a substantial decline in performance, despite achieving ultra-low latency. Thus, there may be a trade-off between service latency and model performance. This approach also supports dynamic expert selection, enabling the system to select any number of experts as required. The process can be iteratively repeated, as outlined in Algorithm \ref{alg:mechan}. Dropping an expert is implemented by assigning a weight of zero to that expert, thereby excluding the corresponding device from participating in the processing.
\begin{algorithm}[t]
	\renewcommand{\algorithmicrequire}{\textbf{Input:}}
	\renewcommand{\algorithmicensure}{\textbf{Output:}}
	\caption{Expert Selection Policy for WDMoE}
	\label{alg:mechan}
	\begin{algorithmic}[1]
		\REQUIRE Weights $\mathbf{w}^i_j$, and latency vectors $\mathbf{t}^i_j$ for experts
		\ENSURE Optimized expert selection set ${\mathbf{Q}}$
		
		\STATE Initialize threshold $\theta \gets 0.5$
		\STATE Select Top-2 experts based on weights ${\mathbf{Q}} \gets \text{Top-K}(\mathbf{w}^i_j, K=2)$
		\STATE Compute initial WLR: $\sum_{k=1}^U \hat{WLR}^i_k$ using Eq.~\eqref{WLR_eq}
		
		\WHILE{$\sum_{i=1}^3 \sum_{k=1}^U WLR^i_k \leq 1.01 \times \sum_{i=1}^3 \sum_{k=1}^U \hat{WLR}^i_k$}
		\STATE Compute cosine similarity $S(\mathbf{w}^i_j, \mathbf{t}^i_j)$ using Eq.~\eqref{sim}
		
		\IF{$S(\mathbf{w}^i_j, \mathbf{t}^i_j) \leq \theta$ \textbf{and} $j \in \mathcal{U}$} 
		\STATE Update $\mathbf{Q}$ by dropping the expert with lower weights for the $j$-th token
		\ENDIF
		
		\STATE Update threshold: $\theta \gets \theta + 0.1$
		\ENDWHILE
		
		\STATE Recalculate cosine similarity $S(\mathbf{w}^i_j, \mathbf{t}^i_j)$
		\STATE Update $\mathbf{Q}$ by removing experts with lower weights for the $j$-th token
		
		\RETURN $\mathbf{Q}$
	\end{algorithmic}
\end{algorithm}

\subsection{Bandwidth Allocation}
Given the expert selection $\mathbf{Q}$, the upper level optimization objective function is reorganized as
\begin{equation}
    \begin{aligned}
        t^i(\mathbf{B}) = & \max_{k \in \mathcal{U}}  \left\{ \sum_{j=1}^{J} q^i_{j,k} \left[ \frac{L^{comm}}{B_{k} \log_2 \left(1 + \frac{P^d_{k} g_{BS,k}}{N_0 B_{k}}\right)} \right. \right. \\
        & \left. \left. + \frac{L^{comm}}{B_{k} \log_2 \left(1 + \frac{P^u_{k} g_{k,BS}}{N_0 B_{k}}\right)} + \frac{L^{comp}}{C_{k}} \right] \right\} , 
    \end{aligned}
    \label{cvx_func}
\end{equation}
and Problem P1 is equivalent to 
\begin{align}
    &\hspace{-1.1cm}\llap{\text{P3:}} & \min_{\mathbf{B}} \quad & \sum_{i=1}^{I} t^i \nonumber \\
    && \text{s.t.} \quad & \sum_{k=1}^{U} B_k = B, \\
    &&& B_k \geq 0, \ \forall k \in \mathcal{U}.
\end{align}

Eq.(\ref{cvx_func}) can be can be viewed as a composite form of several functions:
\begin{gather}
	t^i(\mathbf{B}) = \max_{k \in \mathcal{U}} \{f_k(B_k)\}, \label{max}\\
	f_k({B}_k) = \sum_{j=1}^J q^i_{j,k} g_{j,k}({B}_k) ,\label{sum}\\
	g_{j,k}({B}_k) = h(v^u_{j,k}({B}_k)) + h(v^d_{j,k}({B}_k)) + \frac{L^{comp}}{C_k}, \label{compo}
\end{gather}
where 
\begin{gather}
	h(v) = \frac{L^{comm}}{v} \\
	v^u_{j,k}(B_k) = B_{k} \log_2 \left(1 + \frac{P^u_{k} g_{k, BS}}{N_0 B_{k}}\right), \\
	v^d_{j,k}(B_k) = B_{k} \log_2 \left(1 + \frac{P^d_{k} g_{BS, k}}{N_0 B_{k}}\right).
\end{gather}

The domain of variable $B_k$ is $B_k > 0$ and convex. The first derivative of $v^u_{j,k}(B_k)$ is given by:
\begin{equation}
	\begin{aligned}
		{v^u_{j,k}}^\prime(B_k) = & \log_2\left(1 + \frac{P^u_{k} g_{k, BS}}{N_0 B_{k}}\right) - \\
		& \frac{P^u_{k} g_{k, BS}}{\ln 2} \frac{1}{N_0 B_{k} + P^u_k g_{k, BS}}.
	\end{aligned}
\end{equation}

The second derivative of $v^u_{j,k}(B_k)$ is given by:
\begin{equation}
	\begin{aligned}
		{v^u_{j,k}}^{\prime\prime}(B_k) = \frac{{P}^u_k g_{k ,BS}}{N_0 \ln2} \frac{-\frac{{P}^u_k g_{k ,BS}}{N_0}}{B_k (B_k + \frac{\underline{P}^u_k g_{k ,BS}}{N_0})^2} \leq 0.
	\end{aligned}
\end{equation}

Since ${v^u}^{\prime\prime}_{j,k}(B_k) \leq 0$, we conclude that $v^u_{j,k}(B_k)$ is concave. In a similar way, $v^d_{j,k}(B_k)$ is also a concave function. The first derivative of $h(v)$ is $h^{\prime\prime}(v) = -\frac{L^{comm}}{v^2} < 0$. The second derivative of $h(v)$ is $h^{\prime\prime}(v) = \frac{2 L^{comm}}{v^3}$ and $h^{\prime\prime}(v) > 0$ if $v > 0$. Here $h$ is convex and nonincreasing. According to \cite{boyd2004convex}, $h(v^d_{j,k}({B}_k))$ and $h(v^u_{j,k}({B}_k))$ are convex.

As a result, (\ref{compo}) is a form of nonnegative weighted sums that preserves convexity, and (\ref{sum}) is also a nonnegative weighted sums operation. It is proved that $f_k(B_k),\ k \in \mathcal{U}$ is convex. Therefore, its pointwise maximum $t^i(\mathbf{B})$ is also convex. 

Given a fixed $\mathbf{Q}$, the constraint of P3 is affine, and the domain of $B_k$ is convex. Thus, Problem P3 is convex. It is convenient to solve convex optimization problem through various open source convex optimization algorithm, and Sequential Least Squares Programming implemented in SciPy\cite{2020SciPy-NMeth} is adopted in this paper. 

\addtolength{\topmargin}{0.04in}
\begin{table*}[t]
	\centering
	\caption{Scores of different models on various benchmarks.}
	\label{performance}
	\begin{tabular}{lccccccccccc}
		\hline
		Model & Active Params & MMLU & PIQA & ARC-E & ARC-C & Humaneval & GSM-8K & BoolQ & MBPP \\
		\hline
		Llama 2 7B & 7B & 46.8\% & 78.3\% & 56.1\% & 40.3\% & 12.8\% & 16.7\% & 74.9\% & 14.8\% \\
		Llama 2 13B & 13B & 55\% & 79.8\% & 71.8\% & 60.3\% & 18.9\% & 29.6\% & 82.4\% & 26.8\% \\
		Llama 2 70B & 70B & 69.7\% & 82.5\% & 85.9\% & 78.3\% & 26.2\% & 63.5\% & 87.7\% & \textbf{39.6}\% \\
		Mistral 7B-v0.1 & 7B & 64.1\% & 81.6\% & 83.6\% & 74.2\% & 22.6\% & 47.5\% & 84.1\% & 32.0\% \\
		Mixtral 8x7B-Instruct-v0.1 & 13B & \textbf{70.0}\% & 83.2\% & {92.8}\% & 84.8\% & 47.6\% & 70.9\% & {88.72}\% & 35.2\% \\
		WDMoE& $<$13B & 68.98\% & \textbf{83.51}\% & \textbf{93.12}\% & \textbf{86.78}\% & \textbf{48.17}\% & \textbf{71.29}\% & \textbf{88.87}\% & \textbf{37.4}\% \\
		\hline
	\end{tabular}
\end{table*}


\section{Simulation Results}
In this section, we evaluate the performance of the proposed WDMoE based on numerical results obtained from experiments conducted on NVIDIA A40 GPUs.
\subsection{Experiment Settings}
In the simulation, we consider an MEC server deployed at BS and 8 mobile devices. The expert network $q$ in each MoE layer is deployed on the $q$-th device. The distance between the $q$-th device and BS is denoted as $d_{q}$. We consider Rayleigh fading channels with a mean $10^{-\frac{PL(d_{q})}{20}}$, where the path loss is $PL(d_{q})(dB) = 32.4 + 20log_{10}(f^{carrier}) + 20log_{10}(d_{q})$. The carrier frequency $f^{carrier}$ is set as $3.5$ GHz. The transmission power of BS and the device is 10 Watts and 0.2 Watts, respectively. The total bandwidth is 100 MHz, allocated evenly among all devices. 

We leverage OpenCompass Platform\cite{2023opencompass} to conduct an in-depth and holistic evaluation of large language models. The benchmarks include MMLU\cite{hendrycks2020measuring}, PIQA\cite{bisk2020piqa}, ARC-Easy, ARC-Challenge\cite{clark2018think}, Humaneval\cite{chen2021evaluating}, GSM-8K\cite{cobbe2021training}, BoolQ\cite{clark2019boolq}, MBPP\cite{austin2021program}. 
\begin{figure}[t]
	\centerline{\includegraphics[width=1\columnwidth]{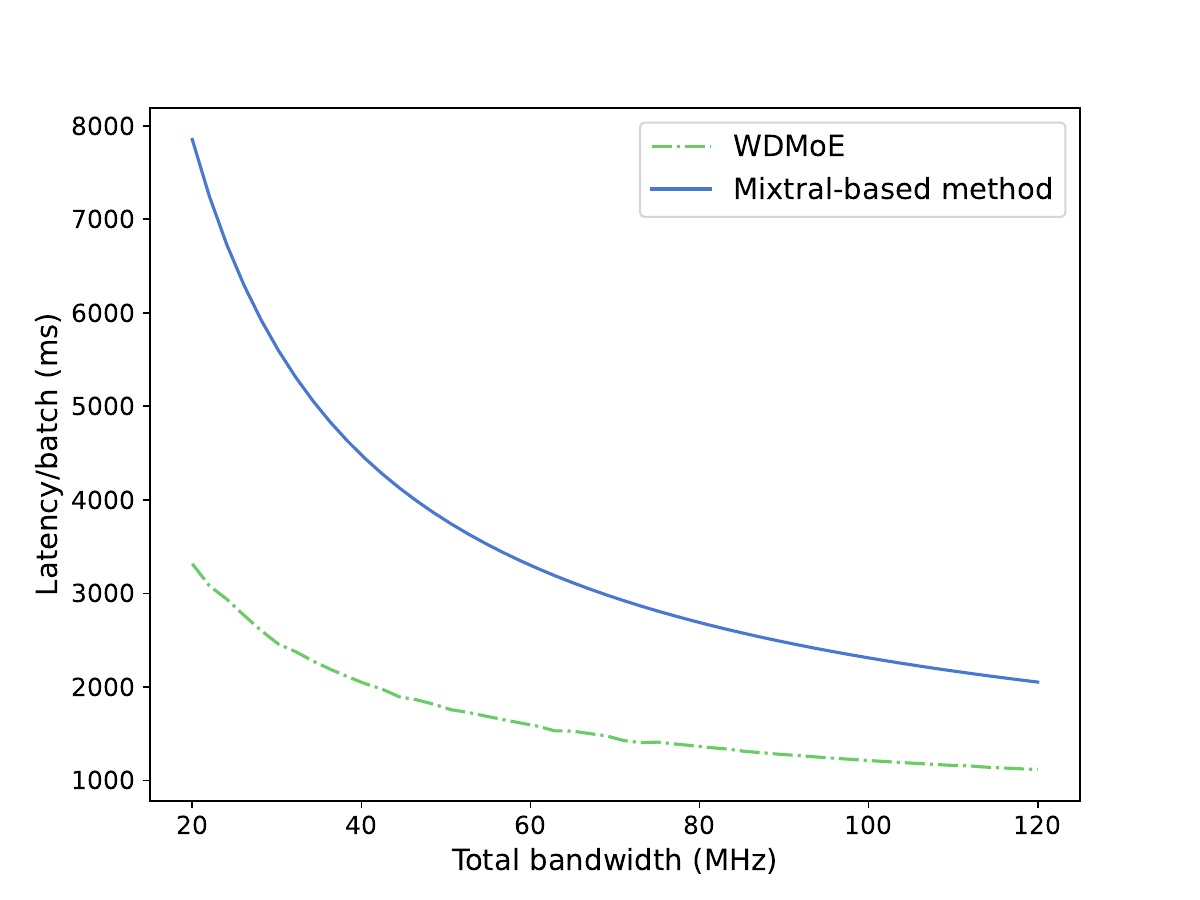}}
	\caption{The latency per batch data versus total bandwidth based on ARC-C dataset.}
	\label{B_optimization}
\end{figure}

\subsection{Performance Evaluation}
We compare WDMoE with state-of-the-art models released during the same period, including Llama 2 with 7B, 13B and 70B parameters, as well as Mistral (referred to as Mixtral 7B-v0.1) and Mixtral (referred to as Mixtral 8x7B-Instruct-v0.1), across various benchmarks. Besides, we evaluate the latency of Mixtral and WDMoE in the wireless network. The Mixtral-based method represents distributedly deploy Mixtral and allocates bandwidth evenly.

\textbf{Model Capability.} 
The detailed results are shown in Table \ref{performance}. Due to the dynamic nature of WDMoE, the total number of active parameters in the MoE remains under 13B, which is fewer than both Llama 2 13B and Mixtral in certain instances. From Table \ref{performance}, we observe that the WDMoE model outperforms Llama 2 70B across most benchmarks while activating at most 20\% of the parameters and generally surpassing Mixtral. Specifically, WDMoE achieves the best results on the PIQA, ARC-E, ARC-C, Humaneval, GSM-8K and BoolQ benchmarks. WDMoE shows minimal performance degradation on MMLU benchmarks. In the ARC-C benchmark, WDMoE exhibits significantly higher accuracy compared to both Llama 2 70B and Mixtral, scoring 86.78\% versus 78.3\% and 84.8\%, respectively.
\begin{figure*}[!t]
	\centering
	\subfloat[]{\includegraphics[scale=0.37]{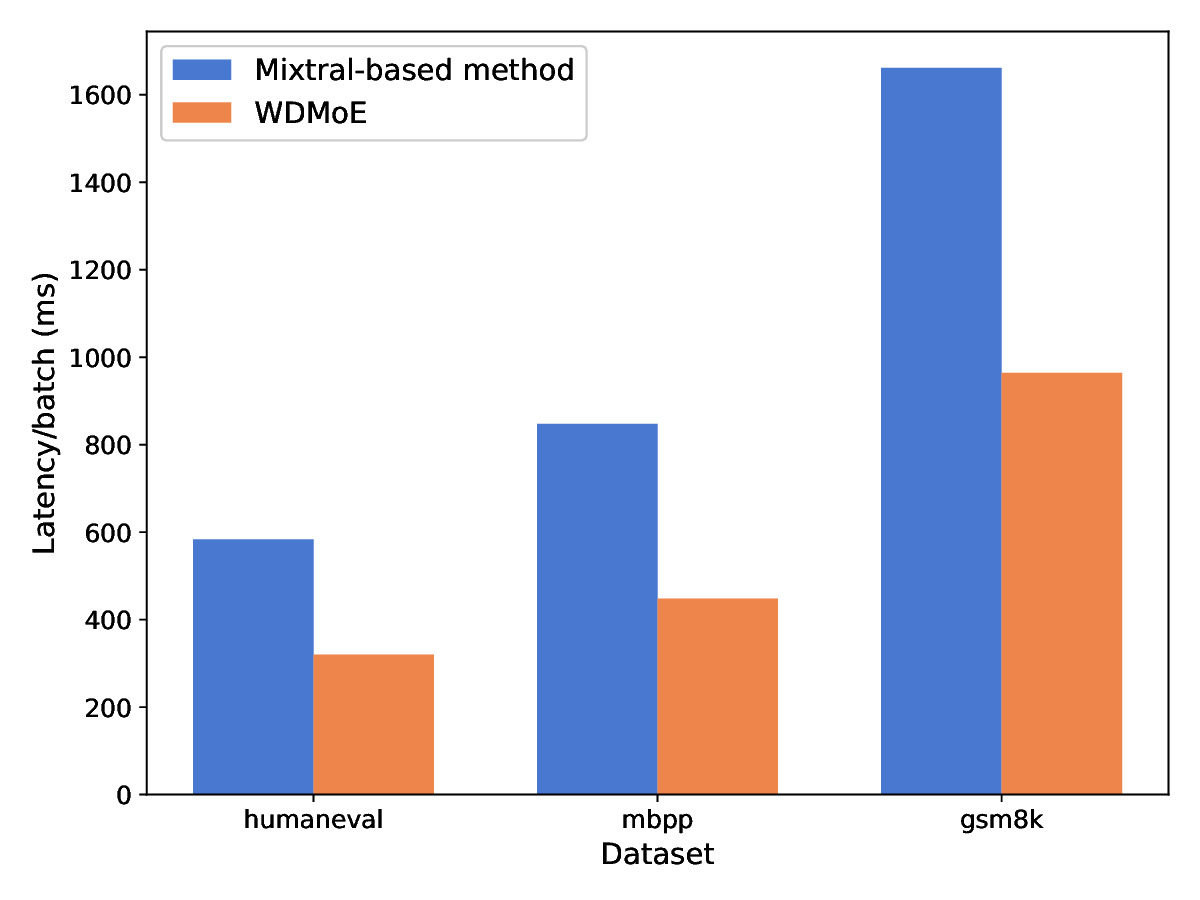}\label{a}}
	\subfloat[]{\includegraphics[scale=0.37]{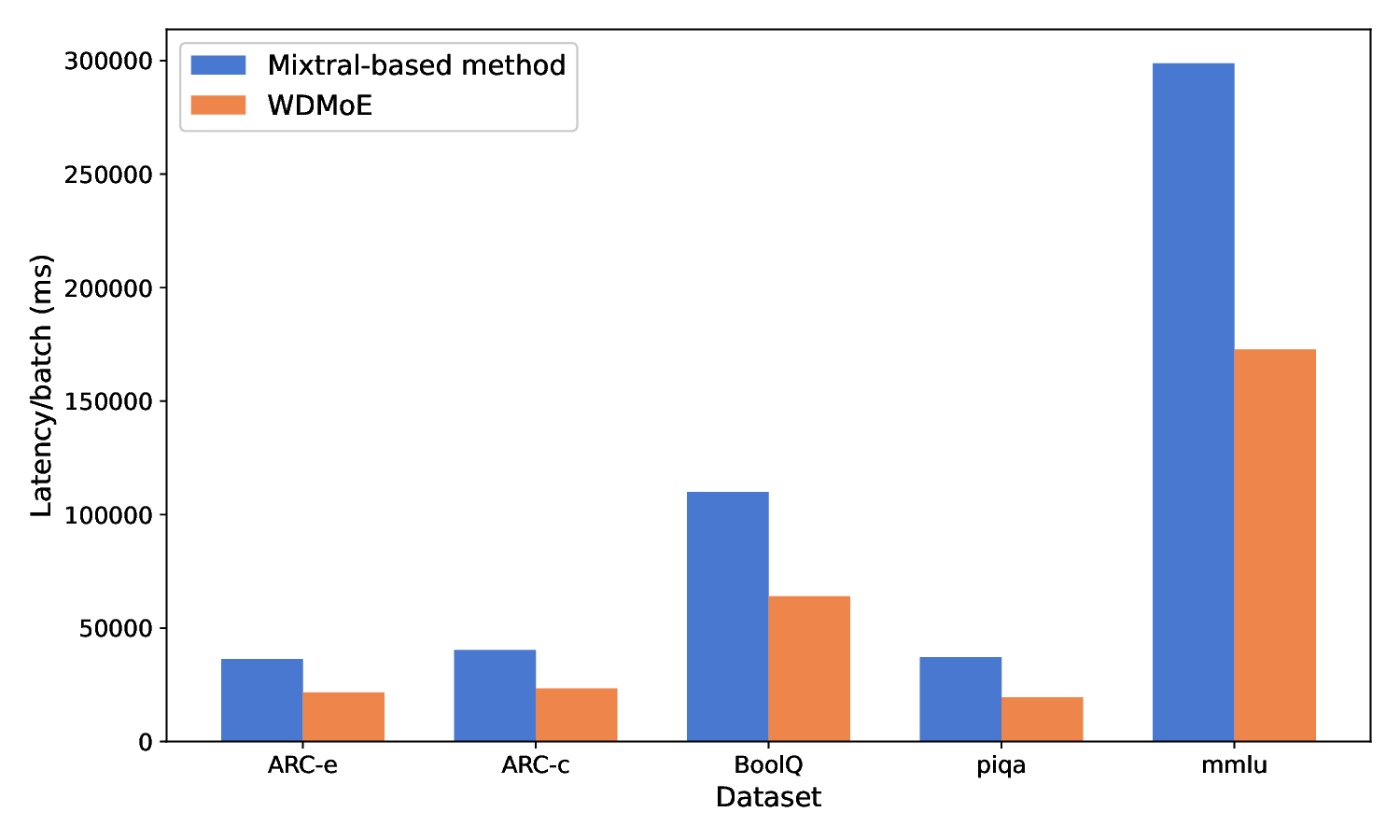}\label{b}}\\
	\caption{Latency versus different datasets: (a) Humaneval, MBPP and GSM-8K datasets; (b) The rest datasets.}
	\label{ab}
\end{figure*}
\begin{figure}[t]
	\centerline{\includegraphics[width=1\columnwidth]{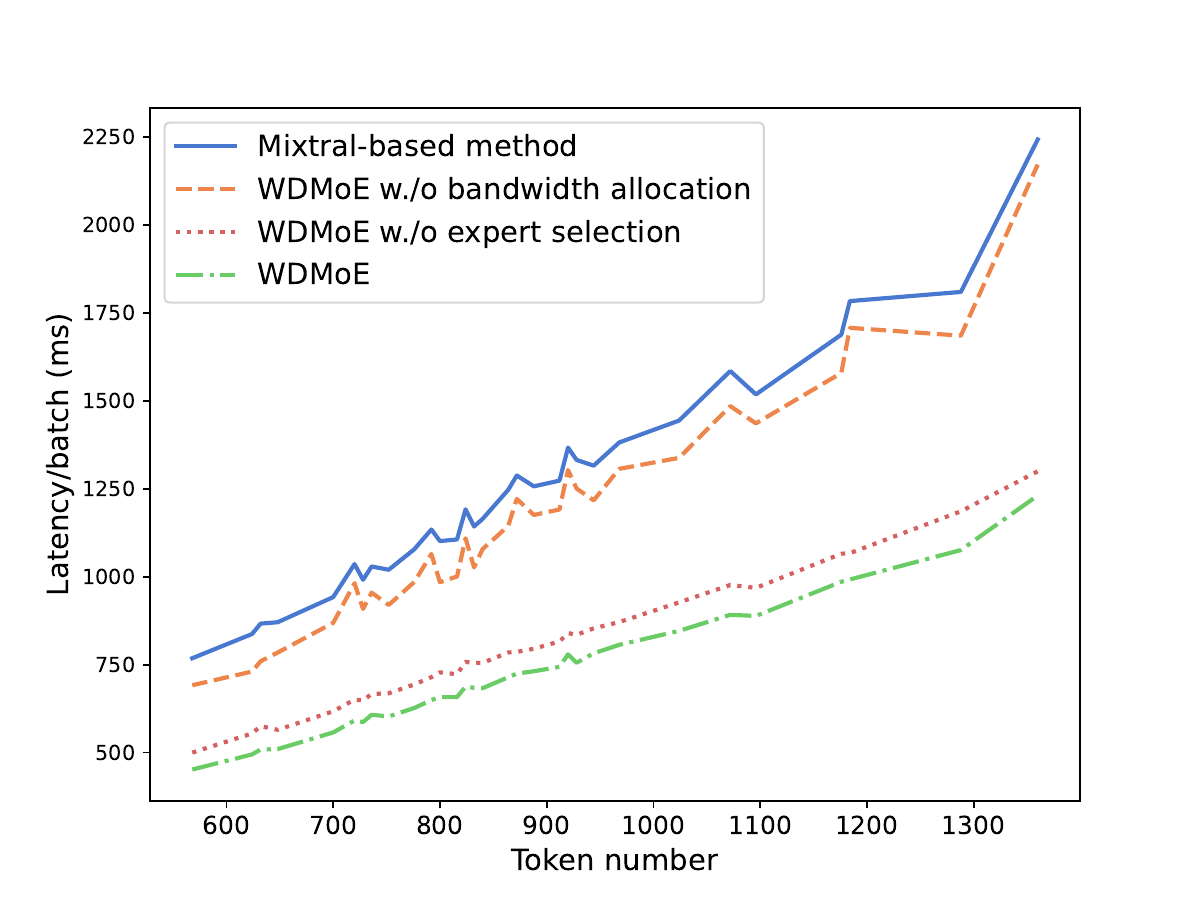}}
	\caption{Ablation study for different cases based on ARC-C dataset.}
	\label{as}
\end{figure}
\textbf{Latency.} We evaluate the end-to-end latency of Mixtral and WDMoE. Fig. \ref{B_optimization} illustrates the latency per batch of data versus total bandwidth. The solid line represents the performance of WDMoE, while the dashed line denotes the performance of Mixtral with the uniform bandwidth allocation. As bandwidth increases, both methods exhibit a decreasing trend in latency. The proposed WDMoE consistently outperforms the Mixtral-based method, by maintaining significantly lower latency across all bandwidth levels. Fig. \ref{ab} presents the average latency per batch with different methods on various datasets. We can see that, compared to the Mixtral-based method, the proposed WDMoE reduces the latency by 41.40\%, 47.14\% and 41.96\% for the corresponding dataset in Fig. \ref{ab}(a) and 40.41\%, 42.03\%, 45.14\%, 47.50\% and 42.19\%  for the corresponding dataset in Fig. \ref{ab}(b).

\subsection{Ablation Study}
We perform ablations on both the upper level and lower level optimizations in the bilevel optimization problem to evaluate their contributions to reducing latency. In Fig. \ref{as}, using the change in latency with the number of tokens from the ARC-C dataset as an example, the expert selection policy achieves a 6.89\% performance gain (WDMoE vs. WDMoE w./o expert selection). Meanwhile, the bandwidth allocation shows a 36.59\% improvement (WDMoE vs. WDMoE w./o bandwidth allocation). The primary reason for this fact is that the upper level optimization focuses on reducing overall attention waiting latency, and its solution is specifically designed to optimize this aspect. The ablation study is conducted on all datasets used in this paper, and the results are presented in Table \ref{abst}. A key finding from Table \ref{abst} is that the expert selection yields a greater improvement when combined with bandwidth optimization compared to the Mixtral with uniform bandwidth allocation method. This suggests that the proposed expert selection is particularly effective in communication scenarios involving bandwidth allocation.

\begin{figure}[t]
	\centerline{\includegraphics[width=1\columnwidth]{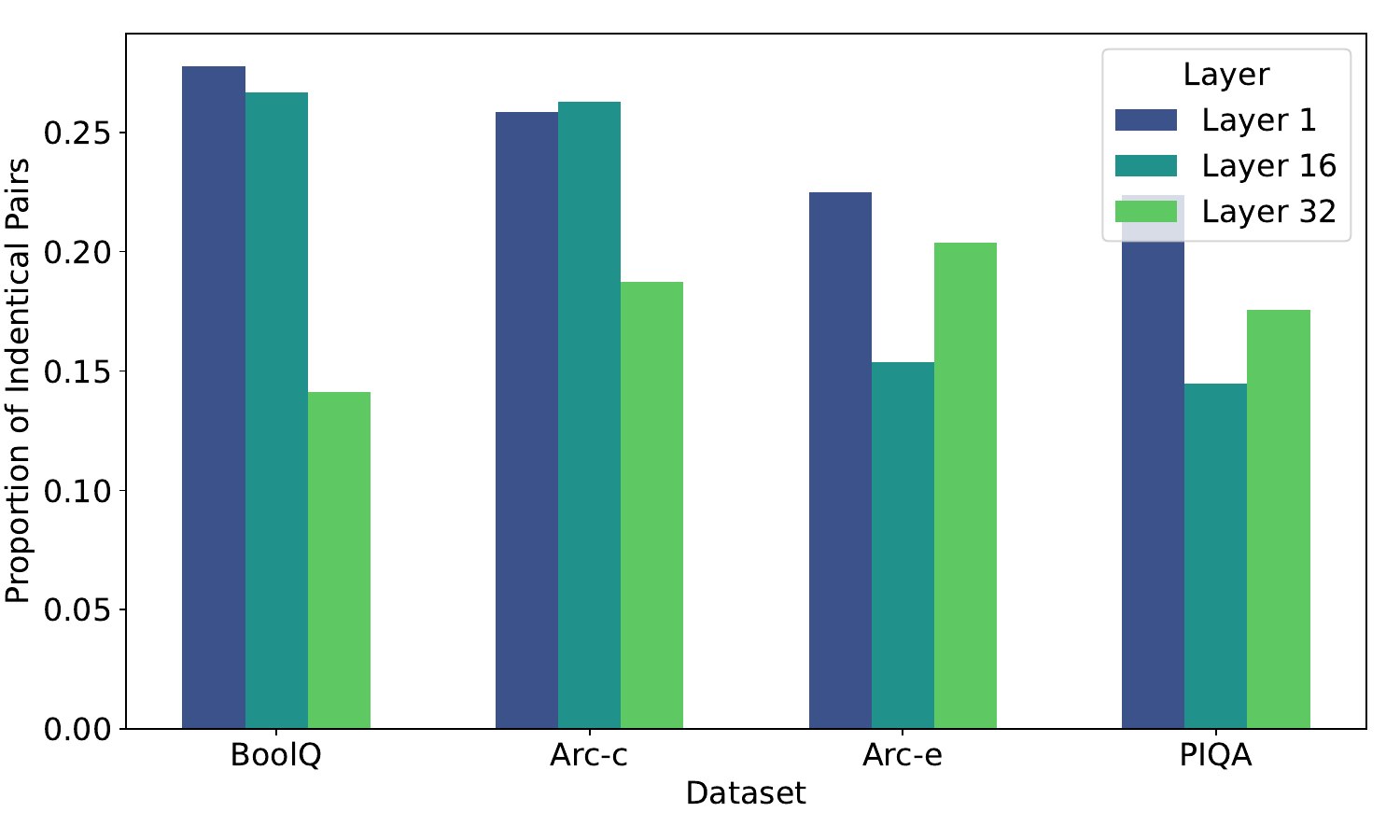}}
	\caption{The maximum ratio of the same expert selection in one batch.}
	\label{Pair}
\end{figure}
\subsection{Insight for Practical Deployment of WDMoE-based LLMs}
In terms of expert network with hundreds of million parameters, as the number of mobile devices in the wireless network increases, mobile devices can load more expert networks from a specific MoE layer while loading fewer expert networks from other MoE layers. In the simulation, we observe a significant phenomenon that various tokens in a text sequence are transmitted to the same experts. Fig. \ref{Pair} shows the proportion of the same expert selection for distinct tokens in the first MoE layer, the sixteenth MoE layer and the thirty-second MoE layer. In the ARC-C dataset, the maximum proportion of expert selection pairs exceeds 25\% in most MoE layers. Similar situations also appeared in other datasets and MoE layers. We can conclude that if we deploy expert networks of an MoE layer using an adjustable method to harness the memory and computing power of a large amount of mobile devices such as deploying the two most frequently selected expert networks for the same token in an MoE layer, the attention waiting latency will be decreased. This is because a more efficient distributed deployment policy can prevent duplicate transmission of a token. Reducing the volume of token transmission at the source will benefit more efficient bandwidth allocation and decrease latency.
\begin{table*}[t] 
	\centering
	\caption{Latency/batch (ms) on all the components of WDMoE.}
	\label{abst}
	\begin{tabular}{lcccccccc}
		\hline
		Components  & MMLU & PIQA & ARC-E & ARC-C & Humaneval & GSM-8K & BoolQ & MBPP \\ 
		\hline
		Mixtral-based Method  & 298813.6 & 37183.1 & 36401.5 & 40367.1 & 572.6 & 1661.6  & 109957.8 & 847.9 \\ 
		WDMoE w./o bandwidth allocation  & 258884.0 & 33861.6 & 35043.3 & 37584.2 & 527.3 & 1491.5  & 106806.9 & 700.9 \\ 
		WDMoE w./o expert selection  & 195383.3 & 22114.1  & 22774.5 & 25598.4  & 335.2 & 1066.0  & 66684.0 & 538.1 \\ 
		WDMoE  & 172743.9 & 19522.2 & 21692.0 & 23400.0 & 305.9 & 964.5  & 63991.0 & 448.2 \\ 
		\hline
	\end{tabular}
\end{table*}
\begin{figure*}[!t]
	\centering
	\subfloat[]{\includegraphics[width=0.8\columnwidth]{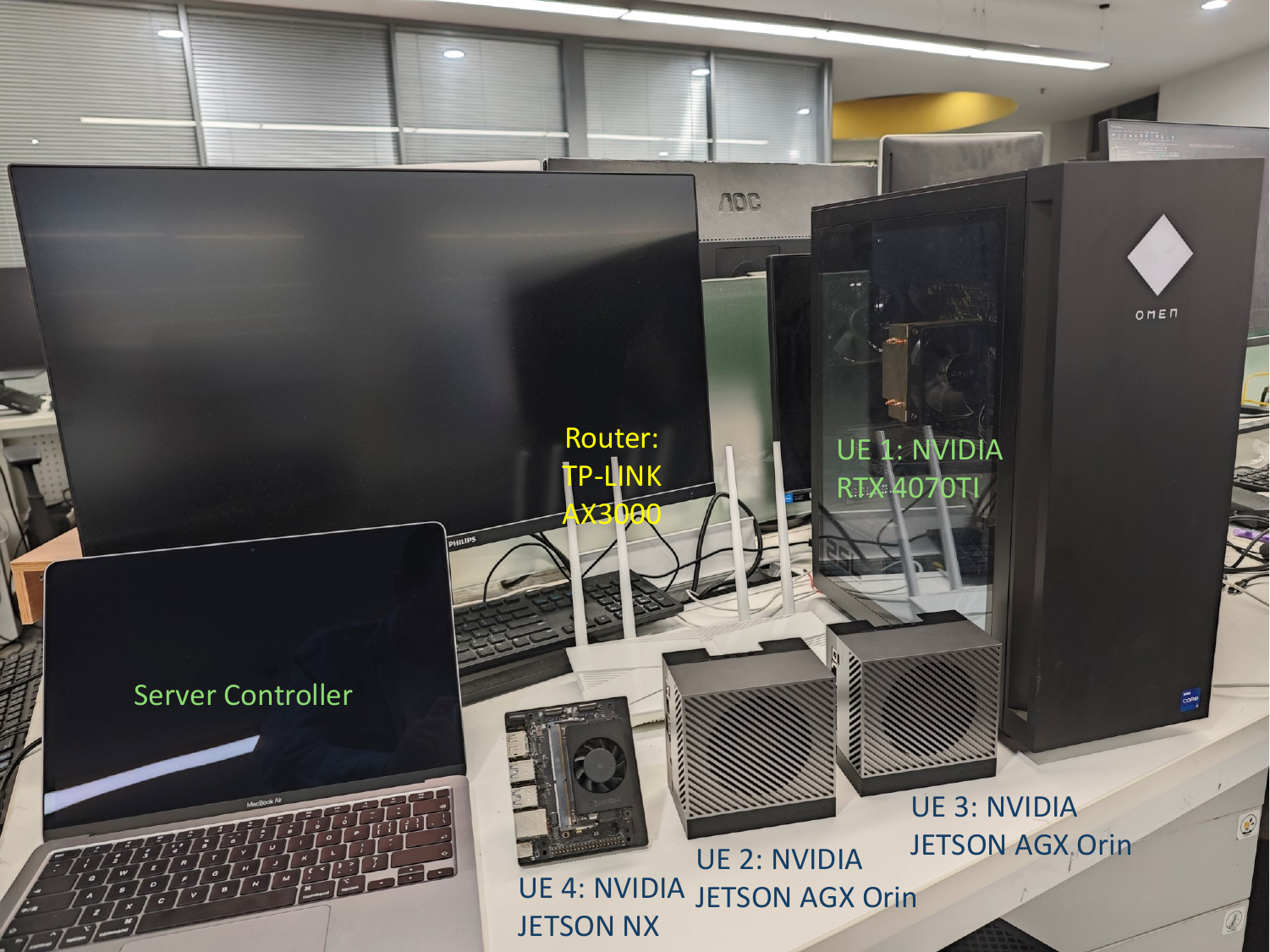}\label{photo}}
	\subfloat[]{\includegraphics[width=1.1\columnwidth]{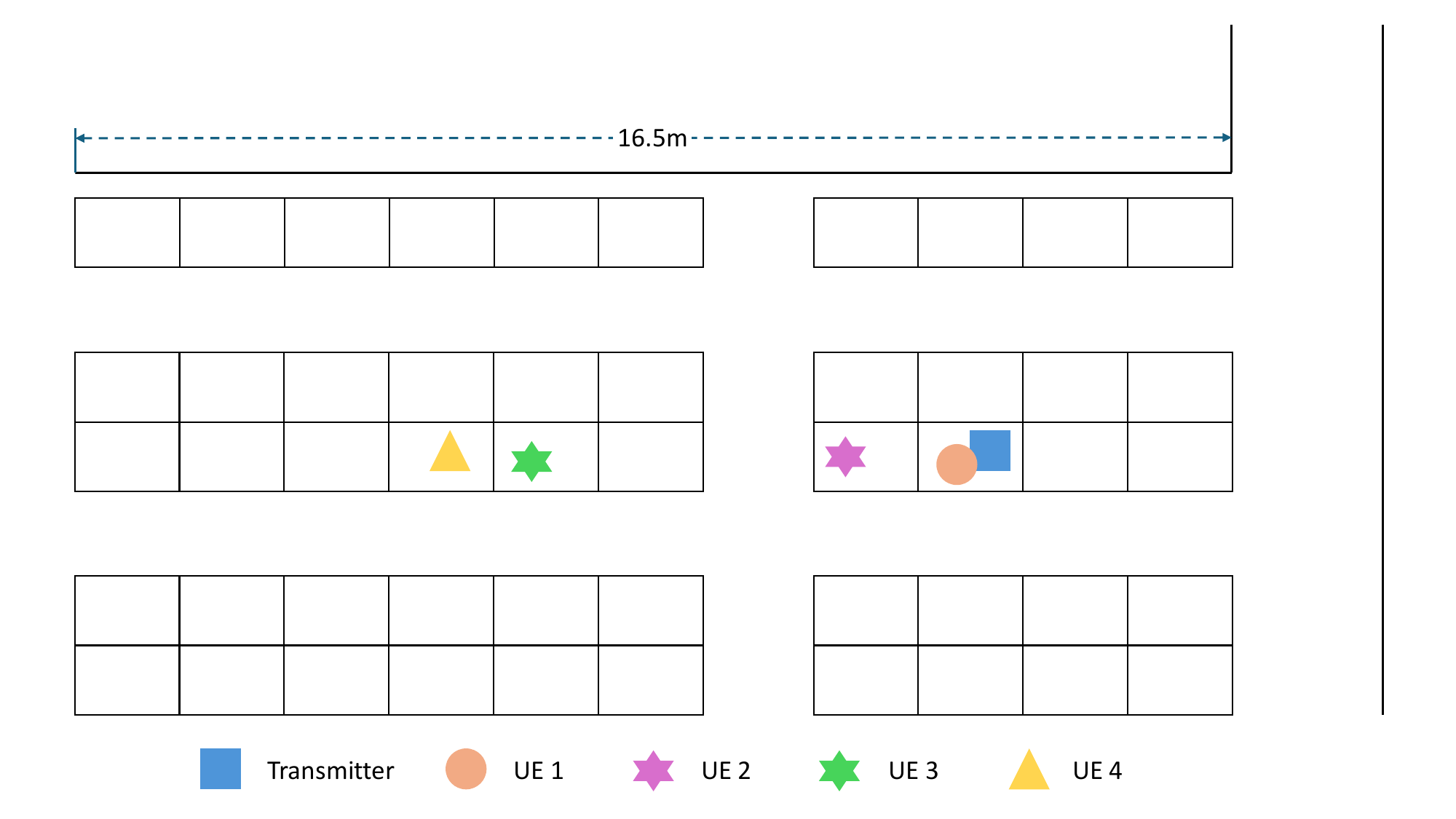}\label{topo}}\\
	\caption{(a) System components of WDMoE hardware testbed; (b) Geographical distribution of experiment devices.}
	\label{hard}
\end{figure*}
\section{Hardware Testbed for WDMoE-based LLMs}
In this section, we develop a hardware testbed for WDMoE using NVIDIA Jetson Kits to demonstrate its efficiency in terms of latency and model accuracy in a practical environment.
\subsection{System Components}
As shown in Fig. \ref{hard}(a), the hardware platform consists of two NVIDIA Jetson AGX Orin devices, an NVIDIA Jetson Xavier NX, a computer with an NVIDIA RTX 4070 Ti, a server equipped with four NVIDIA A40 GPUs, and a TP-LINK AX3000 router.

The router serves as a wireless access point, simulating the BS in our experiment and is responsible for receiving processed results from mobile devices and distributing new tokens to them. Devices equipped with different GPUs exhibit varying computational resources, and communication latency can be controlled by adjusting the physical placement of the devices to vary their distance from the BS.

Each device is equipped with a subset of experts from all MoE blocks. Due to the limited number of devices, only four experts are allocated to each device for each MoE layer, while the remaining expert networks are executed on the server.

In the WDMoE, devices receive tokens from the BS and transmit processed tokens back to it. In our practical experiment, the Transmission Control Protocol (TCP) is used at the transport layer, while the wireless local area network operates based on the IEEE 802.11ax protocol.
\subsection{Simulated Scenarios}
The simulated scenario involves the collaboration of multiple mobile devices for the deployment and inference of LLMs in communication networks, particularly in wireless networks, according to our proposed WDMoE architecture. Fig. \ref{hard}(b) illustrates the geographical distribution. The length of the square is 1.45 m, and the width is 0.8 m. The experiment is conducted indoors. Fig. \ref{hard}(b) shows the layout of the transmitter and mobile devices. The transmitter is centrally located, as represented by the blue square, while the UEs are distributed around it—UE 1 (orange circle), UE 2 and UE 3 (green and purple stars), and UE 4 (yellow triangle). This ensures a realistic wireless network environment for testing.

\subsection{Wireless Distributed Acceleration Algorithm}
The primary purpose of this hardware testbed system is to verify the effectiveness of the WDMoE deployment architecture and the basic concepts for expert selection, without estimating channel conditions, predicting transmission rates, or allocating communication bandwidth to reduce latency. Therefore, the optimization algorithm based on estimated latency, as discussed in Section IV, is not suitable for this system. As analyzed in Section V, balancing the workload across experts can effectively accelerate processing under certain conditions without degrading model performance. Consequently, we propose an expert selection policy specifically designed for this hardware system to achieve faster model inference.

\begin{algorithm}[t]
	\renewcommand{\algorithmicrequire}{\textbf{Input: }}
	\renewcommand{\algorithmicensure}{\textbf{Output: }}
	\caption{Expert Selection for WDMoE Hardware Testbed}
	\label{alg:hard_mechan}
	\begin{algorithmic}[1]
		\REQUIRE Weights $\mathbf{w}^i_j$, and historical latency $\overline{t_k}$
		\ENSURE Selected expert set $\hat{\mathbf{Q}}$
		
		\STATE $\hat{\mathbf{Q}} \leftarrow \text{Top-K}(\mathbf{w}^i_j), K=2$ 
		\STATE Initialize $\hat{J}^k \gets 0, \; k \in \mathcal{U}$ 
		\STATE Initialize $\hat{\mathcal{J}} \gets \{\}, \; \hat{\mathcal{W}} \gets \{\}$
		
		\FOR{$k \in \mathcal{U}$}
		\STATE Calculate the allocated tokens for expert $k$: $J_k$ 
		\STATE Predict latency for expert $k$: $\hat{t}^i_k \gets \text{Eq. \ref{predictt}}$
		\ENDFOR
		
		\STATE Identify latency bottleneck: $\hat{k} \gets \arg\max_{k \in \mathcal{U}} \{ \hat{t}^i_k \}$
		\STATE Analyze upper bound for dropable tokens: $\hat{J}^{drop}_k \gets \text{Eq. \ref{dropnum}}$
		
		\FORALL{$l \in \mathcal{J}$}
		\IF{$w^i_{l,\hat{k}} < \frac{1}{5} \sum_{j=1}^J q^i_{j,\hat{k}} w^i_{j,\hat{k}}$}
		\STATE Add token $l$ to $\hat{\mathcal{J}}$ and its weight $w^i_{l,\hat{k}}$ to $\hat{\mathcal{W}}$
		\STATE Increment token count: $\hat{J}^{\hat{k}} \gets \hat{J}^{\hat{k}} + 1$
		\ENDIF
		\ENDFOR
		
		\IF{$\hat{J}^{\hat{k}} > \hat{J}^{drop}_{\hat{k}}$}
		\STATE Select tokens to drop: $\text{Top-K}(-\hat{\mathcal{W}}), K = \hat{J}^{drop}_{\hat{k}}$
		\STATE Set weights of dropped tokens to zero and update $\hat{\mathbf{Q}}$
		\ELSE 
		\STATE Include tokens in $\hat{\mathcal{J}}$ and update $\hat{\mathbf{Q}}$
		\ENDIF
		
		\RETURN $\hat{\mathbf{Q}}$
	\end{algorithmic}
\end{algorithm}
During program execution, the historical transmission and processing latency for device $k$ at each layer is recorded as $t_k$. The corresponding number of tokens processed by device $k$ is denoted as $J_k$. The average latency per token is given by:
\begin{equation}
	\overline{t_k} = \mathbb{E}\left[\frac{t_k}{J_k}\right].
\end{equation}

Based on experimental data, the variance in latency per token for each expert is sufficiently low, allowing the mean latency per token to serve as a reliable estimate. Since the gating network is deployed on the server, the server can use this average latency per token, along with the number of tokens processed by each device, to predict the total latency for each device:
\begin{equation}
	\hat{t}^i_k = \overline{t_k} \cdot J_k.
	\label{predictt}
\end{equation}

The expert $\hat{k} = \arg\max_{k \in \mathcal{U}} \{{t}^i_k\}$ represents the bottleneck expert, with $\hat{t}^i_{\hat{k}}$ being the estimated time for both communication and computation. When $\hat{t}^i_{\hat{k}}$ is significantly larger than that of other experts, the load assigned to device $\hat{k}$ must be reduced. Specifically, if $\hat{t}^i_{\hat{k}}$ exceeds 1.5 times the third quartile of the predicted latency across all devices, then device $\hat{k}$ is identified as a major contributor to the inference delay of the MoE layer. The third quartile of the predicted latency is denoted as $\hat{t}^i_{\frac{3}{4}}$. The maximum number of tokens that can be offloaded from device $\hat{k}$ is calculated by:
\begin{equation}
	\hat{J}^{drop}_k = \left\lfloor \frac{\hat{t}^i_{\hat{k}} - \hat{t}^i_{\frac{3}{4}}}{\hat{t}^i_k} \right\rfloor
	\label{dropnum}
\end{equation}
where $\lfloor \cdot \rfloor$ denotes round down operation. 

For higher model accuracy, the expert $k$ will be excluded from selection when its weight is excessively low for a specific token. When the weight of expert $k$ for a token is the lowest after Top-K operation and satisfies the condition $w^i_{l,k} < \frac{1}{5} \sum_{j=1}^J q^i_{j,k} w^i_{j,k}$, the token will be dropped and no longer allocated to the expert $k$. If the number of tokens that meet the above conditions exceeds $\hat{J}^{drop}_k$, select the tokens corresponding to the $\hat{J}^{drop}_k$ smallest weights to drop. If the number of tokens that meet the exclusion criteria is less than $\hat{J}^{drop}_k$, the tokens corresponding to $\hat{J}^{drop}_k$ smallest weights will be selected for dropping. If the number of tokens that meet the exclusion criteria is less than $\hat{J}^{drop}_k$, those tokens will not be transmitted to the expert $k$. Otherwise, the tokens with the least $\hat{J}^{drop}_k$ weights will be dropped to avoid the poor performance due to the decrease in total weights. Assuming the latency prediction is accurate, further reducing the number of transmitted tokes will neither lower latency nor maintain model performance. The expert selection process in the hardware testbed system is described in the Algorithm \ref{alg:hard_mechan}. and the expert selection is conducted on the server. The server determines the assignment of token computation through expert selection $\hat{\mathbf{Q}}$, and the router distributes tokens to the devices in the network. 

This method is specifically designed to reduce attention waiting latency by identifying latency bottlenecks and employing targeted expert selection. Its key advantages lie in its simplicity and efficiency, rendering it a viable solution for performance optimization.
\begin{figure}[!t]
	\centerline{\includegraphics[width=1\columnwidth]{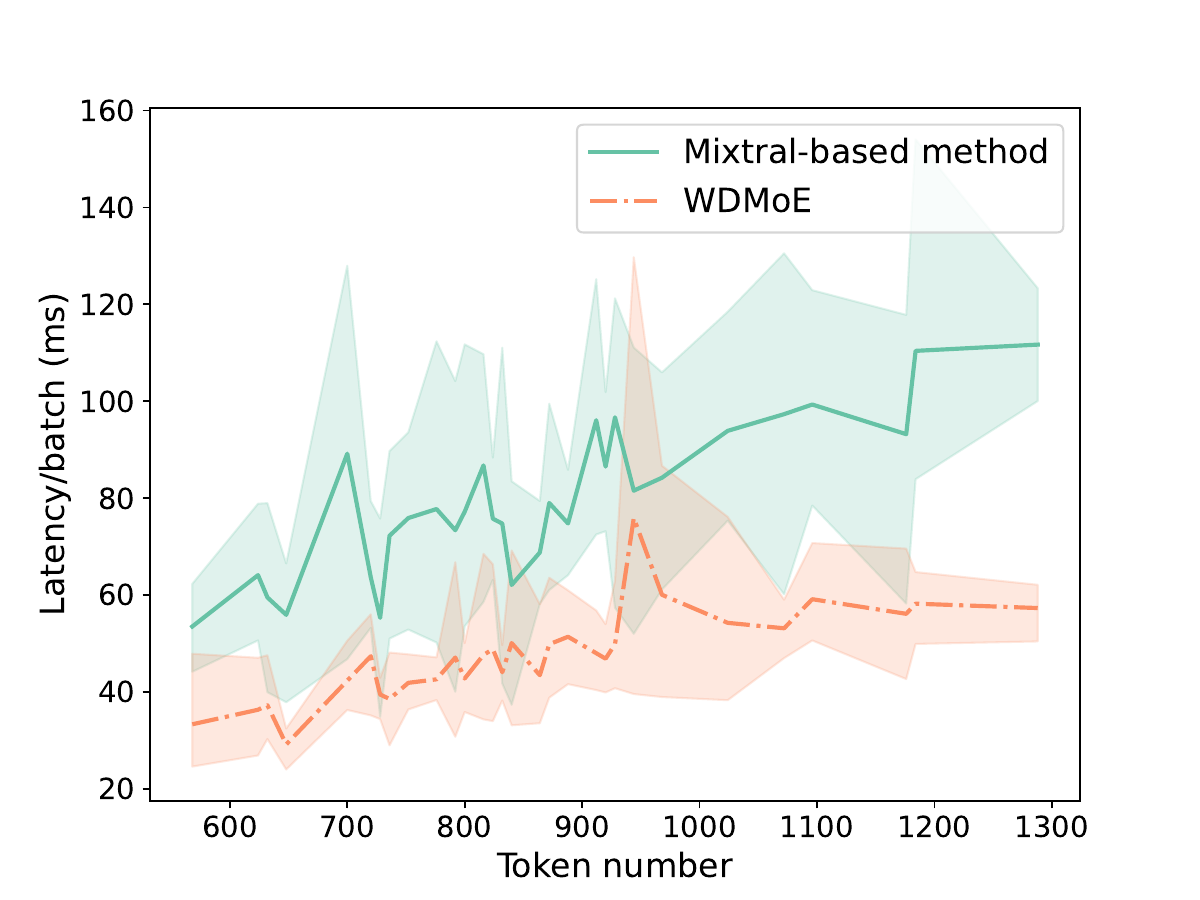}}
	\caption{Latency of different methods on the hardware testbed system.}
	\label{ht}
\end{figure}

\subsection{Hardware Experiment Results}
We evaluate the performance of our hardware testbed system from both model capability and attention waiting latency, based on theoretically simulated experiments, and compare the proposed method with the baseline Mixtral-based method. The Mixtral-based method refers to the Top-K selection method of Mixtral. We conduct three experiments on various benchmarks for both the vanilla selection and our proposed method under the same environmental settings to eliminate random errors to the best of our ability.

Based on theoretical simulated experiment results, the model capability of the hardware system's algorithm is verified on only four benchmarks for convenience, including ARC-E, ARC-C, MBPP, and PIQA. The evaluation results indicate that our algorithm's distributed deployment with expert selection outperforms the vanilla expert selection which does not consider latency. The latency per batch in a layer versus the number of tokens is shown in Fig. \ref{ht}. The line represents the mean latency, while the shaded area indicates the range of latency values. The orange area is generally below the green area, except at two points where the maximum latency values are higher due to channel variation in practical experiments. To present the results more fully, we record the latency data of three groups of comparative experiments in Table \ref{hl}, where the maximum average latency reduction reaches 45.75\%. The hardware system experiments validate our theoretical analysis and experimental results.
\begin{table}[t]
	\centering
	\caption{Model accuracy.}
	\begin{tabular}{lcccc}
		\hline
		Model             & ARC-E & ARC-C & MBPP & PIQA \\ \hline
		Mixtral          & 92.42 & 86.1  & 37.8 & 83.41 \\ 
		WDMoE-testbed    & \textbf{92.95} & \textbf{87.12} & \textbf{38.8} & \textbf{83.51} \\ \hline
	\end{tabular}
\end{table}

\begin{table}[t]
	\centering
	\caption{Latency/batch (ms) in hardware testbed experiments.}
	\label{hl}
	\begin{tabular}{lcccc}
		\hline
		Model             & ARC-E & ARC-C & MBPP & PIQA \\ \hline
		Mixtral-based method-1          & 532.8 & 1625  & 38.77 & 616.7 \\ 
		WDMoE-testbed-1          & 468.3 & 1228  & 37.96 & 414.3 \\
		Mixtral-based method-2          & 418.1 & 2583  & 33.47 & 1380 \\ 
		WDMoE-testbed-2          & 372.6 & 1530  & 29.49 & 436.9 \\ 
		Mixtral-based method-3          & 383.5 & 1406  & 30.72 & 519.4 \\ 
		WDMoE-testbed-3          & 361.9 & 656.6  & 28.33 & 332.0 \\ \hline
		Average Gain    & \textbf{9.536\%} & \textbf{39.523\%
		} & \textbf{7.246\%} & \textbf{45.750\%} \\ \hline
	\end{tabular}
\end{table}

\section{Conclusion}
In this paper, we introduce WDMoE, a wireless distributed MoE architecture for LLMs. This architecture enables the collaborative deployment of LLMs across the MEC server at the BS and mobile devices in the wireless networks. By leveraging the parallel characteristics of expert networks, the deployment architecture effectively models attention waiting latency and incorporates the WLR metric, which jointly assesses model performance and service latency. Through the development of expert selection and bandwidth allocation strategies, we solve the formulated bilevel optimization problem. Besides, we build a hardware testbed to validate the effectiveness of the proposed method. Extensive experiments demonstrate WDMoE ensures high performance and significantly reduces latency, which verifies the feasibility of distributed LLMs in wireless scenarios and highlights the promising future of cooperative edge-device large models. 
\bibliographystyle{IEEEtran}
\bibliography{references}

\vfill

\end{document}